\documentclass{nature}

\usepackage[english]{babel}
\usepackage[T1]{fontenc}
\usepackage{csquotes}
\usepackage[fleqn]{amsmath}
\usepackage{amsmath}
\usepackage{pdfpages}
\usepackage{wasysym}
\usepackage{amsmath}
\usepackage[version=4]{mhchem}
\usepackage{array}
\usepackage{graphicx}
\usepackage[labelfont=bf]{caption}
\makeatletter
\let\saved@includegraphics\includegraphics
\AtBeginDocument{\let\includegraphics\saved@includegraphics}
\renewenvironment*{figure}{\@float{figure}}{\end@float}
\makeatother

\usepackage{xcolor,colortbl}

\definecolor{orange}{HTML}{FF7F00} 
\definecolor{commercial}{HTML}{000000} 
\definecolor{14}{HTML}{004cff} 
\definecolor{15}{HTML}{00ff37} 
\definecolor{19}{HTML}{e600ff} 
\definecolor{20}{HTML}{00fffb} 
\definecolor{21}{HTML}{0a6310} 
\definecolor{22}{HTML}{FF0000} 
\definecolor{23}{HTML}{e8c5e6} 

\newcolumntype{a}{>{\columncolor{Gray}}c}
\newcolumntype{b}{>{\columncolor{white}}c}

\usepackage[labelfont=bf]{caption}

\title{Autonomous optimization of nonaqueous battery electrolytes via robotic experimentation and machine learning}

\begin{document}

\author{Adarsh Dave$^{1,3}$, Jared Mitchell$^{2,3}$, Sven Burke$^{2,3}$, Hongyi Lin$^{1,3}$, Jay Whitacre$^{2,3,*}$, Venkatasubramanian Viswanathan$^{1,3,*}$}

\maketitle

\begin{affiliations}
 \item Department of Mechanical Engineering, Carnegie Mellon University, Pittsburgh, Pennsylvania, 15213, USA.
 \item Department of Materials Science and Engineering, Carnegie Mellon University, Pittsburgh, Pennsylvania, 15213, USA.
 \item Wilton E. Scott Institute for Energy Innovation, Carnegie Mellon University, 
 Pittsburgh, Pennsylvania, 15213, USA.
 \item[*] Lead Contacts: whitacre@andrew.cmu.edu (J. W.), venkvis@cmu.edu (V. V.)
\end{affiliations}


\maketitle

\section*{Summary}
New battery technology will be crucial to the electrification of transportation and aviation\cite{sripad_performance_2017,bills_performance_2020}, but battery innovations can take years to deliver. For battery electrolytes, the many design variables present in selecting multiple solvents, salts, and their relative ratios\cite{ma_study_2017,liu_effects_2017,hall_exploring_2018,zhang_synergistic_2014,weber_long_2019} mean that  optimization studies are slow and laborious, even those restricted to small search spaces. The key challenge is to lower the number and time-cost of experiments needed to formulate an electrolyte for a given objective. 

In this work, we introduce a novel workflow that couples robotics to machine-learning for efficient optimization of a non-aqueous battery electrolyte. A custom-built automated experiment named ``Clio'' is coupled to Dragonfly - a Bayesian optimization-based experiment planner. Clio autonomously optimizes electrolyte conductivity over a single-salt, ternary solvent design space. Using this workflow, we identify 6 fast-charging electrolytes in 2 work-days and 42 experiments (compared with 60 days using exhaustive search of the 1000 possible candidates, or 6 days assuming only 10\% of candidates are evaluated). Our method finds the highest reported conductivity electrolyte in a design space heavily explored by previous literature, converging on a high-conductivity mixture that demonstrates subtle electrolyte chemical physics.

To close the device gap, we test the performance of six conductivity-optimized electrolytes in pouch cells, revealing improved fast charge capability in all discovered electrolytes against a baseline electrolyte selected \textit{a-priori} from the design space. Our closed-loop workflow realizes the promise of autonomous platforms accelerating material optimization.

\section{Introduction}
High-performance batteries are crucial to the electrification of transportation and aviation\cite{sripad_performance_2017,bills_performance_2020}. However, new battery designs can require extensive manual testing for material optimization, which can take years. Designing a material is fundamentally a complex function that takes material formulation as input and outputs performance. Efficient optimization of such a black-box function via machine-learning has been successfully demonstrated in many engineering domains, including catalytic materials\cite{burger_mobile_2020,zhong_accelerated_2020}, photovoltaics\cite{Sun2019,langner_beyond_2020}, solid-state materials\cite{xue_accelerated_2016,kusne_--fly_2020}, and battery charging protocols\cite{attia_closed-loop_2020}.

There is a great deal of research interest in coupling automated experiments to these machine-learning methods\cite{macleod_self-driving_2020,eyke_toward_2021,mistry_how_2021}. The hope is that ``closed-loop’’ approaches display the following traits when compared to traditional design of materials via manual experimentation: 1) closed-loop experiments are able to discover optimal material designs within a given design space; 2) closed-loop experiments discover optima faster and with fewer experiments; 3) closed-loop experiments offer a principled basis for design-of-experiments (DOE), balancing exploiting design regions likely to have optimal performance with exploring regions of unknown performance. These traits have been demonstrated in related fields\cite{xue_accelerated_2016,macleod_self-driving_2020,langner_beyond_2020,burger_mobile_2020,kusne_--fly_2020,Shields2021}, but have not yet been demonstrated in battery material design outside of aqueous electrolytes\cite{dave_autonomous_2020}.

Out of the materials present in a battery, liquid electrolytes are a particular challenge to optimize. There are many choices for solvent\cite{ma_study_2017,liu_effects_2017,hall_exploring_2018} or salt\cite{zhang_synergistic_2014,weber_long_2019}, each potentially yielding vastly different performance; good electrolytes often contain more than three or four components. Both species choice and relative proportions of species matter, creating a high-dimensional design space spanning both good and bad battery performance.

In this work, we develop a robotic platform named ``Clio’’, capable of closed-loop optimization of battery electrolytes. Clio enables high-throughput experiments characterizing transport properties of various blends of non-aqueous solvents and salts. When connected to an experiment planner (a machine learning model in this work), Clio can autonomously explore and optimize an objective over a given design space. 

Electrolytes can be optimized for different applications. The electrolyte design often must fulfill multiple, competing objectives within each application\cite{aurbach_design_2004}, so optimal designs for rate-capability can differ from optimal designs for cycle-life. In this work, we consider optimization for fast-charging. Fast-charging electrolytes must be able to transport lithium to the anode at high rates, which is strongly associated with bulk transport properties (conductivity, viscosity, diffusivity, cation transference number) and anode interfacial kinetics (charge transfer impedance, desolvation dynamics)\cite{gao_methodologies_2022,logan_electrolyte_2020}. We focus initially on single objective optimization of the bulk conductivity as an objective for improving rate-capability. While this is an incomplete objective function, the workflow introduced in this paper can also enable effective multi-objective optimization\cite{paria_flexible_2020} of electrolytes in future studies.

In this work, Clio autonomously optimizes conductivity over solvent mass fraction and salt molality in a EC-DMC-EMC-LiPF6 ternary solvent, single salt system. Optimal electrolytes then pass through a sequence of fast-charging tests conducted in pouch cells. These results are reported against a baseline electrolyte selected a-priori from the design space.

\section{Automated Electrolyte Characterization}

Clio is illustrated in Figure 1, and was custom-built for autonomous operations by the authors. Given an electrolyte electrolyte, Clio will dose the formula from pre-made feeder solutions. Clio measures the conductivity, viscosity, UV-vis spectrum, and density of the specified electrolyte. The system is rinsed between each composition tested, and all experiments run in triplicate as developed during a contamination study (Extended Data Figure 1); the mean absolute difference between repeated measurements of the conductivity of the same electrolyte is found to be $\pm 1.3\%$ across a range of electrolytes (Extended Data Figure 2 and 3). Samples can also be retained for easy use in follow-up testing in pouch cells. 

Clio is controlled over HTTP - design of experiments are passed to Clio in JSON, and experimental results are returned, enabling integration with any experiment planner. Details on Clio's hardware, software, and operation are contained in Methods.

\begin{figure}
    \centering
    \includegraphics[width=0.75\textwidth]{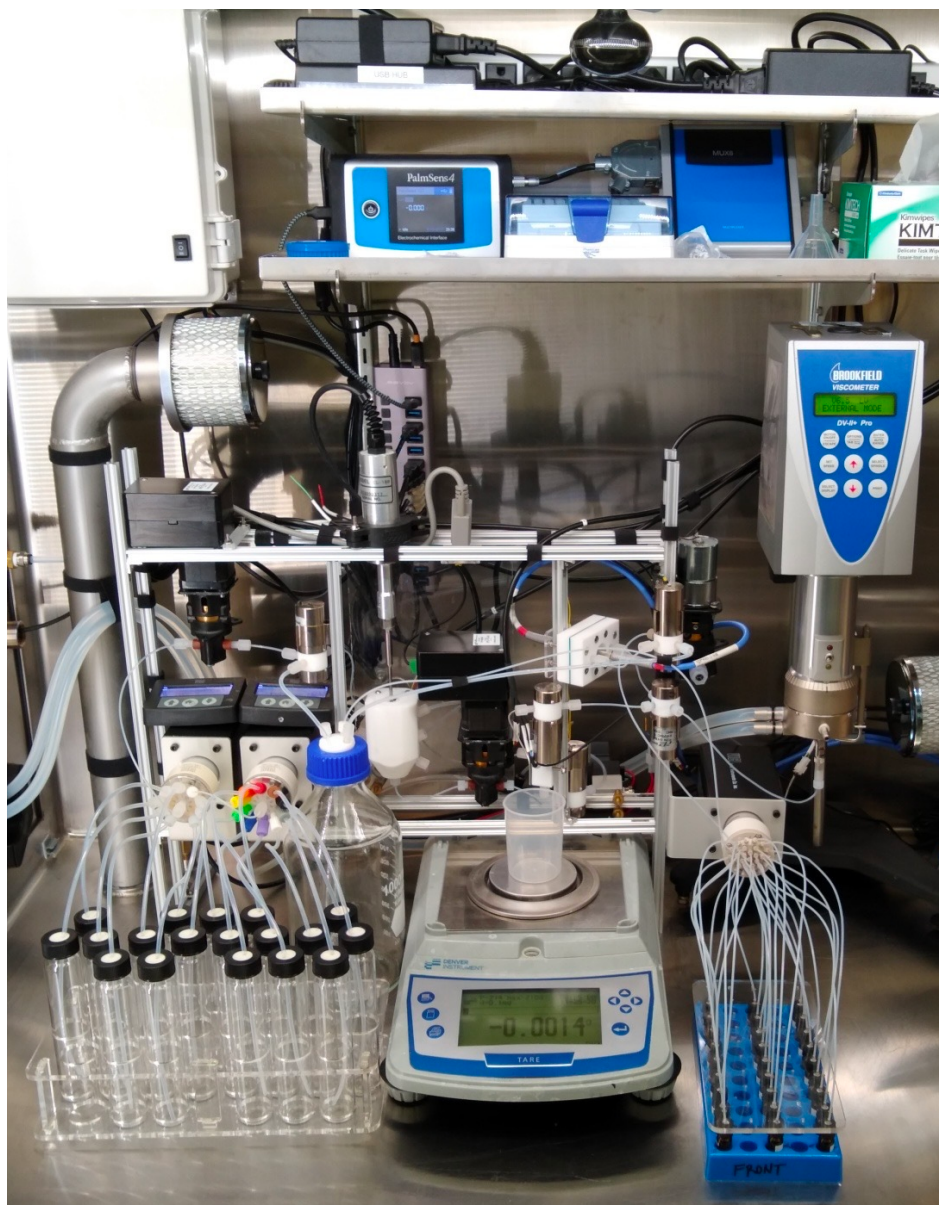}
    \caption{Image of Clio, the robotic nonaqueous electrolyte experiment. Samples are created by mixing up to 19 feeder solutions. Measurements are taken of conductivity, density, viscosity, and UV-vis spectrum. Clio can retains tested samples for follow-on cell testing, enabling rapid screening of electrolytes from property measurement to cell performance.
}
    \label{fig:my_label}
\end{figure}

\section{Machine-Learning Guided Experimentation}
Dragonfly is an open-source Bayesian optimization software package that uses a suite of acquisition strategies and evolutionary algorithms for scalable, robust black-box optimization\cite{kandasamy_tuning_2020,paria_flexible_2020}. Dragonfly’s ease-of-use “out-of-the-box”, adaptive sampling strategies, and broad support for discretized and constrained domains were of interest to the authors; thus, Dragonfly was used as the experiment planner for this work.

A system of ternary solvents featuring ethylene carbonate (EC), ethyl-methyl carbonate (EMC), and dimethyl carbonate (DMC) with lithium hexafluorophosphate ($\ce{LiPF6}$) as salt was chosen for study. This is a well-known space for electrolyte design and an appealing domain to demonstrate a novel experimental method. EC-EMC 30\%-70\% mass fraction with 1.1 mol $\ce{LiPF6}$ per kilogram solvent was chosen a-priori as a baseline for the conductivity optimization and fast-charging target. The temperature of the sample under test was taken during each conductivity evaluation, and remained between 26 \textdegree C and 28 \textdegree C for all measurements reported.

The domain was represented to Dragonfly by three axes: 1) EC mass fraction, 2) DMC co-solvent ratio, calculated as (mass fraction of DMC) / (1 - mass fraction of EC), and 3) $\ce{LiPF6}$ molality. Axis 1 was limited to EC mass fractions of 30\% to 50\% and axis 2 was limited between 0 and 2 mols LiPF6 per kilogram solvent, keeping with conventional choices for electrolyte design. Each axis was split into 10 to 12 equivalently spaced levels, creating more than 1000 points within the domain to search.

Dragonfly optimized for conductivity over 40 experiments in this domain. The electrolytes investigated are illustrated as points in the 3-dimensional design space in Figures 2 and 3. Dragonfly builds an adaptive DOE via four different acquisition functions evaluated over a Gaussian process regression-based surrogate model - Thompson sampling, expected-improvement and top-two expected improvement, upper confidence bound. These four sampling algorithms roughly balance exploration and exploitation. We interspersed the optimization run with periodic random sampling to further favor exploration (Figure 2). This is a heuristic choice, but a rigorous consideration of such hyper-parameters will feature in future work.

\begin{figure}
    \centering
    \includegraphics[width=\textwidth,clip, trim = {0 0 0 0}]{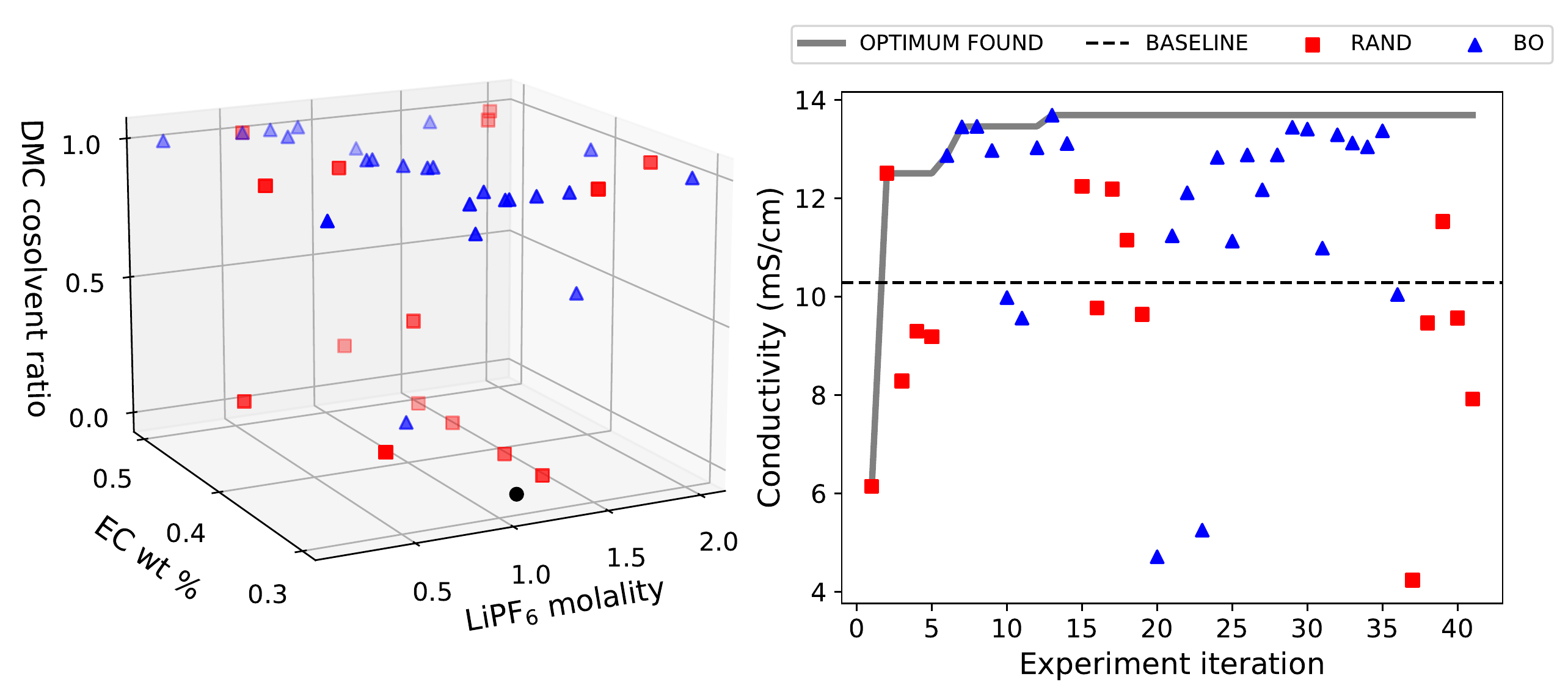}
    \caption{\textit{Left}: electrolytes sampled by Clio during autonomous optimization of conductivity in an EC-EMC-DMC, $\ce{LiPF6}$ design space. Black circle indicates baseline electrolyte selected a-priori (EC/EMC 30/70 by mass, 1.1m LiPF6) \textit{Right}: learning rate of Clio during this optimization. Random sampling (RAND) is interspersed with Bayesian optimization (BO) to bias the optimization towards exploring the space. Dashed black line indicates conductivity of baseline electrolyte as measured by Clio during survey pre-optimization.}
    \label{fig:my_label}
\end{figure}

Figure 2 illustrates a convergence on a conductivity optimum within 15 experiments. The optimizer heavily evaluates 1) the high conductivity, high DMC region compared to the lower conductivity, high EMC region and 2) the middle band of salt concentration between 0.7 to 1.3 mols LiPF6 per kilogram solvent, known to be near the concentration of peak conductivity in standard nonaqueous electrolyte systems. Samples from the high and low salt, mixed DMC-EMC regions could be desired and could be added in future work with alternative DOEs.

\begin{figure}
    \centering
    \includegraphics[width=\textwidth,clip, trim = {0 0 0 0}]{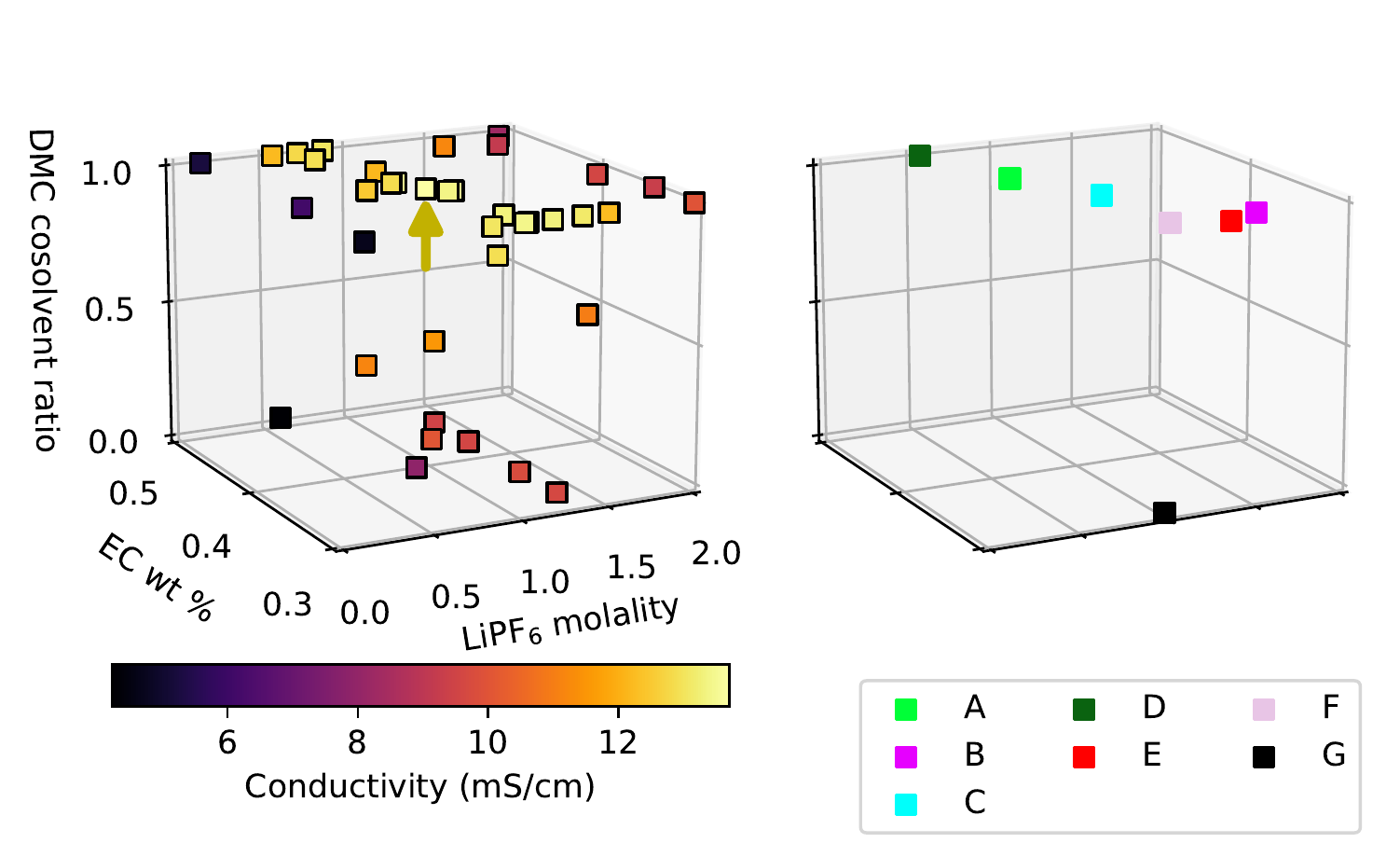}
    \caption{\textit{Left}: measured conductivities (in color) of electrolytes sampled by Clio during autonomous optimization of conductivity in an EC-EMC-DMC, $\ce{LiPF6}$ design space. The gold arrow indicates the highest conductivity electrolyte found (approximately EC:DMC 40:60, 0.9m LiPF6) \textit{Right}: electrolyte candidates chosen for testing in pouch cells. Blends A-F are discovered by Clio. Blend G is the baseline electrolyte.}
    \label{fig:my_label}
\end{figure}

Figure 3 shows measured conductivities in the three-dimensional design space. Historically, linear carbonates like DMC are mixed with cyclic carbonates like EC to lower viscosity and increase conductivity\cite{aurbach_design_2004}, but recent studies reveal that conductivity can also increase with higher dielectric constant as it improves ion dissociation (e.g. EC over DMC, DMC over EMC) \cite{logan_study_2018}. In our study, the conductivity optimum of the EC:DMC:EMC \ce{LiPF6} system at 26-28 \textdegree C is found at EC:DMC 40:60 by mass, 0.9m \ce{LiPF6} with 13.7 mS cm$^{-1}$ (gold arrow in Figure 3). Further experiments were conducted on the EC:DMC face of the design space along the 30\%, 40\%, and 50\% by EC mass contours, confirming this optimum in this design space (Extended Figure 6). Theoretical calculations conducted with the Advanced Electrolyte Model (a highly accurate model for nonaqueous electrolyte transport properties\cite{dave_benchmarking_2019,logan_ester_study_2018}) also show a higher conductivity in the 40\% EC blend compared to 30\% or 50\% - this could be due to the improved ion dissociation in the 40\% compared to the 30\%, and lowered viscosity in the 40\% relative to the 50\% (Extended Figure 7). Thus, while the EC:DMC:EMC \ce{LiPF6} system is heavily explored in literature, our combination of machine-learning and automation has enabled measurement of the highest conductivity blend yet reported from this space.

Out of 42 evaluations, seven candidate electrolytes were picked heuristically for follow-up cell testing. Three of the highest conductivity electrolytes were chosen (blends A, C, and F in Figure 3), as were two higher-salt concentration electrolytes with >10 mS/cm (blends B and E) and one lower-salt concentration electrolyte with >10 mS/cm (blend D). As mentioned above, EC-EMC 30\%-70\% by weight, $\ce{LiPF6}$ 1$m$ was a baseline electrolyte denoted blend G. The cell testing candidates are given with their color-codes in Figure 3 and in Table 1.

\begin{table}
\begin{tabular}{|c|r|c|c|c|}
  \hline
  Name & Color-code & Solvent Mass Fractions & LiPF6 Molality & Conductivity (mS/cm) \\
  \hline
  G &\cellcolor{commercial} & EC-EMC  30\%-70\%   & 1.1 & 9.8 \\
  \hline
  A &\cellcolor{15} & DMC-EC-EMC 45\%-50\%-5\%  & 1.1 & 12.2\\
  \hline
  B & \cellcolor{19} & DMC-EC 70\%-30\%  & 1.5 & 12.1\\
  \hline
  C & \cellcolor{20} & DMC-EC 60\%-40\%  & 0.9 & 12.8\\
  \hline
  D & \cellcolor{21} & DMC-EC 50\%-50\%  & 0.5 & 10.8\\
  \hline 
  E & \cellcolor{22} & DMC-EC 70\%-30\%  & 1.3 & 12.1\\
  \hline 
  F & \cellcolor{23} & DMC-EC 70\%-30\%  & 1.0 & 12.4\\
  \hline
\end{tabular}
\caption{Table of electrolyte compositions involved in autonomous optimization. Blend C and F had highest measured conductivities in the space; blend B and E had high salt concentrations and >10 mS/cm conductivity; blend D had a low salt concentration and >10 mS/cm. Blend A had highest conductivity with EMC present. Blend G was chosen as a baseline electrolyte \textit{a priori}}
\end{table}

\section{Pouch Cell Testing of Discovered Electrolytes}
A suite of tests on small pouch cells was devised to test electrolyte candidate for fast-charging capability. Candidate electrolytes were mixed by Clio and routed into the retain rack (blue tray, bottom left of Figure 1). Dry pouch cells of 220 mAh capacity were received from the manufacturer with electrode materials, current collectors, and tabs pre-made. A researcher emptied Clio's retain rack and manually injected each electrolyte into these dry pouch cells.

Cells were subjected to formation cycles, then to a five step rate test of increasing constant-current charges up to 4C, with 0.5C constant-current discharges between each step. Cells were then repeatedly cycled at 4C charge, 0.5C discharge until failure due either to over-voltage (>4.3V) or capacity fade. Further details on cell testing protocol are given in Methods.

\begin{figure}
    \centering
    \includegraphics[width=\textwidth,clip]{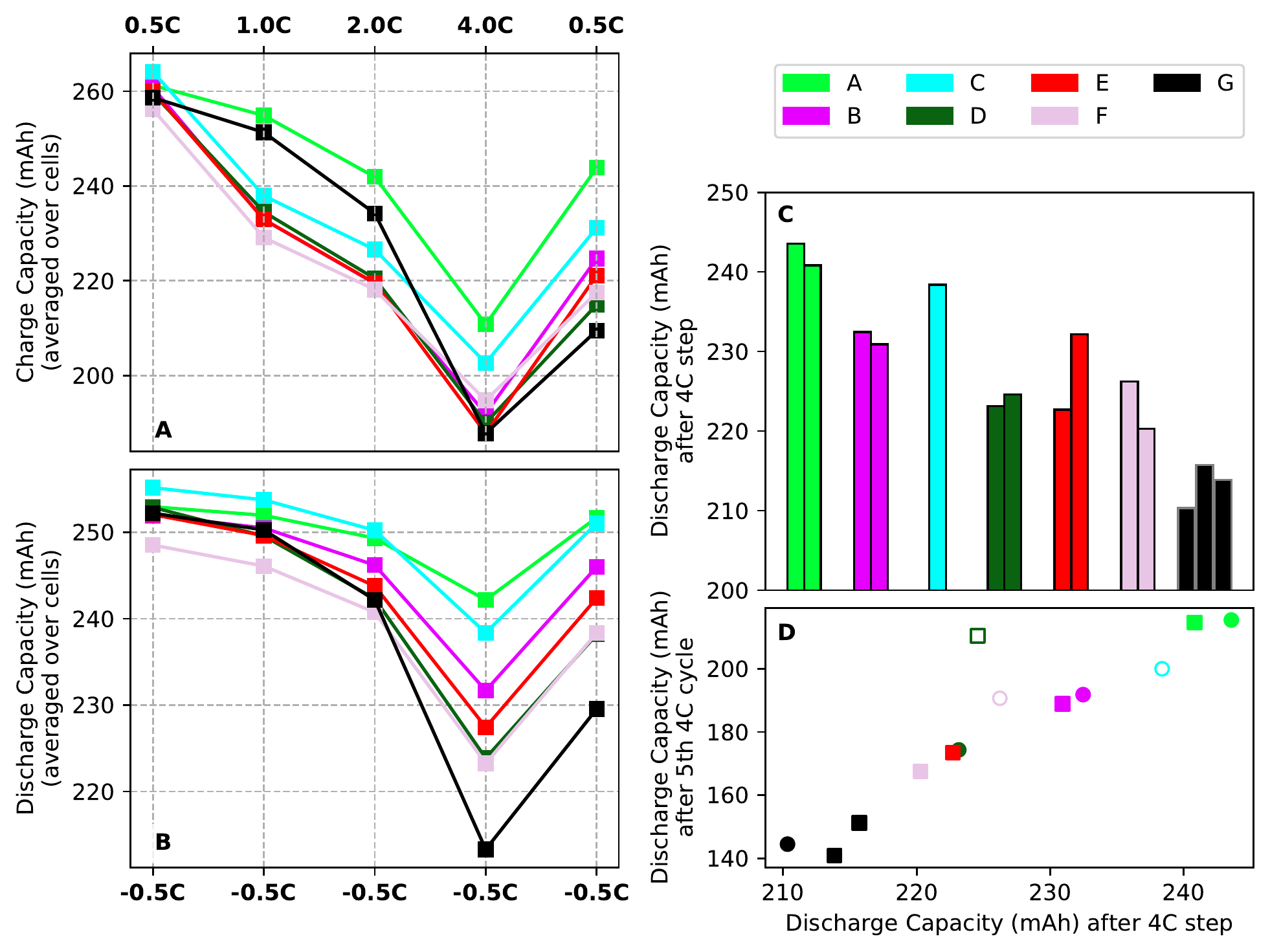}
    \caption{Performance of various electrolytes during rate test. \textit{A}: constant-current charge capacity of each electrolyte, averaged from two cells. \textit{B} : constant-current charge capacity of each electrolyte, averaged from two cells. \textit{C}: bar chart of each cell's discharge capacity after 4C charge. \textit{D}: relationship between performance at 4C rate test step and capacity of cell after 5 back-to-back 4C charge cycles; unfilled markers are cells that did not make it 5 cycles - the last cycle capacity is reported here.}
    \label{fig:my_label}
\end{figure}

Figures 4(A-C) give the results from the rate-test for each candidate electrolyte. Electrolyte blends optimized by Clio all show greater or equal discharge capacity at the 4C step compared to baseline, showing greater usable capacity put into the cells at this high charging rate (Figure 4B). The worst cell containing a Clio electrolyte showed a 5\% improvement on discharge capacity after 4C charging compared to the worst baseline cell, and the best cell with a Clio electrolyte showed a 13\% improvement on this metric compared to the best baseline cell (Figure 4C). Capacity fade during extended 4C charge cycles, along with coulombic efficiency and related metrics, are shown in Extended Data Figure 3 and 4. While the electrolytes in this study are not optimized for long cycle life, discharge capacity after the 4C step of the rate-test shows a strong correlation with discharge capacity after five cycles of 4C charging (Figure 4D).

These results indicate that our novel, closed-loop workflow is able to discover optimal material designs within a given design space, and additionally delivered a conductivity optimum not yet reported in literature despite a heavily studied design space. We note that the baseline electrolyte in this study was picked on intuition - near 1 mol salt per solvent kilogram, relatively low EC content, and commercial availability. The optimization delivered better fast-charging performance in a pouch cell compared to the baseline after only 40 experimental evaluations optimizing for bulk conductivity.

\section{Assessing Workflow Efficiency}
Our closed-loop workflow would be particularly useful if optimal materials are discovered faster or with fewer experiments compared to manual design. Efficiency in optimization is split into two contributing factors: 
time efficiency and sample efficiency. Time efficiency asks the following question: given 40 experiments, is Clio or an equivalent graduate student faster? Sample efficiency asks: given an optimization target, will Clio or an equivalent graduate student achieve it with fewer samples? Given these two estimates, the overall efficiency will combine time and sample efficiency into a question: given an optimization target, will Clio or an equivalent graduate student achieve it faster?

\begin{figure}
    \centering
    \includegraphics[width=\textwidth,clip]{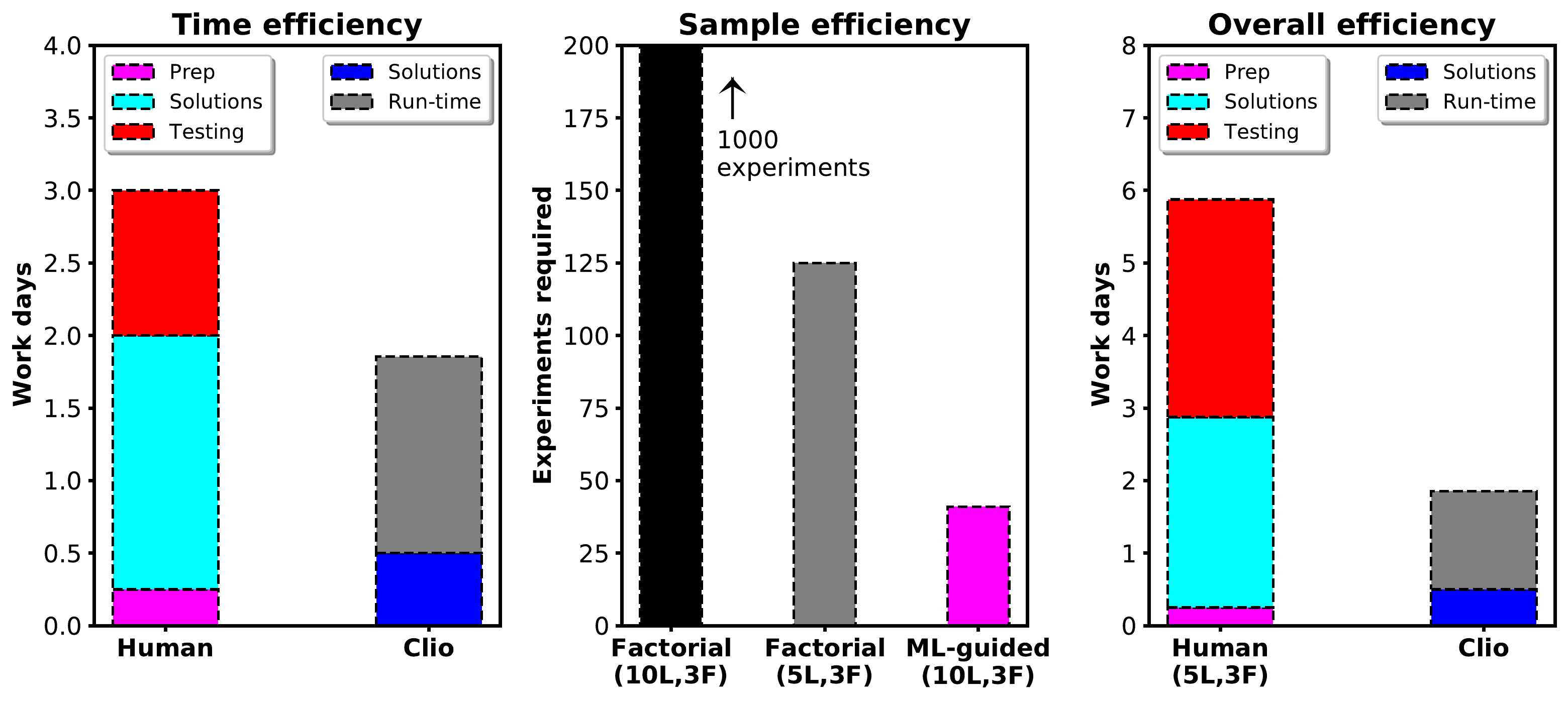}
    \caption{Evaluation of Clio's efficiency compared to human-led design. \textit{Left}: time efficiency of Clio compared to human-led experiments for a fixed design-of-experiments. \textit{Center}: sample efficiency of Clio compared to a full factorial DOE for a fixed objective. \textit{Right}: overall speed-up of Clio compared to human-led experiments for a given objective}
    \label{fig:my_label}
\end{figure}

One co-author documented their time spent running sequential electrolyte characterizations for a previous paper\cite{dave_autonomous_2020}, much in the way Clio operates (given in Extended Data). Based on their estimates accounting for prep time, solution making time, and manual testing time, Clio requires approximately 33\% fewer work-days to complete 40 experiments than a graduate student. Robust software and hardware for automation has enabled Clio to run 24 hours a day, compared to 8 hours a day for a researcher (Figure 5). Sample efficiency is estimated by comparing a factorial DOE with Clio’s ML-guided DOE. This would overestimate samples needed as human DOEs are more efficient than a full factorial. To correct for this, we compare Clio to a human-run, 5-level, 3-factor factorial (representing 10\% of the design space posed to Clio). This still requires 3x the samples than Clio’s DOE which was performed at a 10-level, 3-factor discretization (Figure 5). The overall efficiency considering time and sample efficiency is approximately 3x - Clio will perform in 2 work-days what would take a graduate student greater than one work-week.



\section{Conclusion}
We demonstrate a novel, ``closed-loop'' optimization of a nonaqueous battery electrolyte for conductivity and then close the device gap to show improvements in performance of a pouch cell.

Our workflow using a Bayesian experiment planner produces a highly efficient DOE, finds a yet-unreported conductivity optimum in a well-studied design space, and reveals candidates with better fast-charging performance in a cell than an intuitively chosen baseline. This demonstrates the potential of closed-loop experiments to discover optimal material designs within well-explored and unexplored design spaces. 

By comparing the time and sample efficiency of this scheme to a manually conducted optimization, we estimate a 3x overall speed-up using our scheme. We believe this work will be useful beyond the battery community; our custom-designed robotic platform, experiment planning, and integration with device testing will be valuable in optimizing other autonomous discovery platforms for energy and drug discovery.

\section*{Methods}
\subsection{Clio Hardware}

The electrolyte sample under test is enclosed in chemical resistant material throughout dose, mix, and measurement phases; because a shared, closed volume is encountered by each sequential sample, a regular rinse cycle is required between samples. Acetonitrile is used as the rinse solvent in this work due to its low viscosity, quick evaporation and ability to dissolve LiPF6. The contamination study in Extended Data Figure 1 
shows that, after a rinse, two experiments will converge on the known conductivity of a standard. All experimental evaluations are run in triplicate for this reason, with the final two runs in each triple averaged and reported as the measurement. Repeatability of the conductivity measurement is assessed in Extended Data Figure 2 and 3; 120 samples were taken in triplicate across a range of carbonate solvents (EC, DMC, EMC, DEC) and LiPF6 concentrations; these were assessed for the variance between run 2 and run 3 in each set, finding a mean averaged percent error of $1.3\%$ and a 95\% confidence interval of $\pm 3.8\%$.

The temperature of the sample under test was taken during each conductivity evaluation, and remained between 26 \textdegree C and 28 \textdegree C for all measurements reported.

Conductivity is measured by impedance spectroscopy, run within a custom-built PTFE fixture in which the liquid sample fills a cell with symmetric platinum electrodes. Complex impedance is measured at five frequencies between 14 kHz and 800 kHz - the real part of the impedance at the frequency with the smallest measured phase difference is taken as the resistance of the sample. A cell constant is derived from single point calibration to an acetonitrile solution of known conductivity (see Extended Data Figure 1) - inverse resistance divided by cell constant is reported as the specific conductivity of the sample.

Components used in Clio include a Fluid Metering Inc high precision, stepper motor pump for dosing; a Vivi Valco motorized, programmable valve-set; and a Palmsens4 for running impedance spectroscopy.

Clio is kept in a glovebox maintaining a dry argon atmosphere, with moisture levels measured every 5 minutes to be <10 ppm.

\subsection{Clio Software}

Clio is orchestrated with a suite of Labview tools developed in-house, focusing on low-level device communication and timings. Labview manages a webhook for supplying Clio a specific DOE; electrolyte compositions are passed to the webhook via HTTP, and Clio’s measurements are passed back in the response. Experiment and inventory management is handled through a Python API that interfaces with this webhook; any generic machine-learning recommender can interface with this given appropriate configurations. 

Because Clio mixes by volume, conversion code is needed to transform Dragonfly’s axes to volumes of arbitrary feeder solutions. Doing so requires a-priori estimates of feeder solution density; these values were estimated via the Advanced Electrolyte Model\cite{gering_prediction_2017,dave_benchmarking_2019}, a high-fidelity electrolyte calculation software, and confirmed by Clio’s density measurement.

The client and database sits on an AWS EC2 instance with Toyota Research Institute.

\subsection{Materials Availability}

The solvents and solutes used in this investigation, except for the baseline electrolyte and acetonitrile, were obtained as battery-grade materials from GELON LIB Group. Materials used for pouch-cell testing were also obtained from the GELON LIB Group. Baseline electrolyte of 1.1m LiPF6 in EC:EMC 30:70, was obtained from Sigma-Aldrich. Anhydrous acetonitrile was also obtained from Sigma-Aldrich.

Stock feeder solutions were made for this investigation through first making solute-free mixtures of solvents, then gradually adding the appropriate mass of solute. All solutions were mixed for a minimum of 30 minutes past the dissolution of the last visible solute. All measurements of solvent and solute were done by mass using a Thermofischer brand microbalance. All solutions were mixed with a VWR brand magnetic stir bar and magnetic stir plate. All glassware and stir bars were washed thoroughly with acetonitrile between solutions and were allowed to dry completely before any more solutions were made. Miscibility and co-solubility screening tests were conducted on these feeder solutions before solutions were used in the test stand. Solutions were routinely checked for stability of dissolved species and were routinely inverted to prevent any stratification of the solutions. Feeder solutions were stored in 60 ml amber glass vials with Sure/Seal septa lids. All materials were stored and handled in a dry argon atmosphere with <5ppm moisture. All experimental procedures, except cell testing, were also conducted in a dry argon atmosphere. 

\subsection{Cell Testing}

Dry pouch cells with a nominal capacity of 220mAh were sourced from Linyi Gelon LIB Co. Ltd.. Cells contained \ce{LiNi_{0.5}Mn_{0.3}Co_{0.2}O_{2}} cathode and graphite anode. Cells were run in duplicate or triplicate - blend C was only run once due to a cell presenting with incorrect capacity.

All tests were completed on Neware BTS-CT-4008-5V12A battery cyclers. Cells were cut open and dried in a vacuum oven overnight and then transferred into a glove box. Cells were then filled with 0.6 mL of electrolytes by an Eppendorf pipette and sealed with a heat sealer. Finally, cells were transferred out for the following four steps done in sequence: 1) Cell construction followed by a 36 hour rest held at 1.5V for electrolyte infiltration, 2) Cell formation at C/20 symmetric for three cycles, 3) Rate-test with charge rates of C/2, 1C, 2C 4C, then C/2 again, discharging each time at C/2. Each constant-current charge was completed with a constant-voltage hold at 4.2V until C/20. Each C/2 discharge was completed with a constant-voltage hold at 2.5V until C/20.

Then, each cell was subjected to a cycling test until failure - either over-voltage of 4.3V during the charge or low capacity. Each cycle had a charge rate of 4C, discharge rate of 0.5C. completing each cycle again with a CV hold at 4.2V. Each C/2 discharge was completed with a constant-voltage hold at 2.5V. Cells were rested for 10 minutes between each step of the test.

\subsection{Lead Contact}
J. Whitacre and V. Viswanathan are reachable at whitacre@andrew.cmu.edu, and venkvis@cmu.edu respectively.

\begin{addendum}
\item This work was supported by Toyota Research Institute through the Accelerated Materials Design and Discovery program.  The authors acknowledge insightful discussions with Brian Storey, Abraham Anapolsky, Linda Hung and Chirranjeevi Gopal from the Toyota Research Institute and Biswajit Paria from Carnegie Mellon University.
\item[Contributions] 
\item[Corresponding Authors] Correspondence and requests for materials should be addressed to Jay Whitacre, ~(email: whitacre@andrew.cmu.edu) and Venkat Viswanathan, ~(email: venkvis@cmu.edu).
\item[Declaration of Interests] The authors declare that they have no competing financial interests.
\item[Data Availability] All data generated in this work will be deposited into $matr.io$ platform and publically available. Additionally, the datasets generated during and/or analysed during the current study are available from the corresponding author on reasonable request.
\item[Code Availability] The code associated with the work can be made openly available upon request.

\end{addendum}

\bibliographystyle{naturemag}
\bibliography{nature_bib}

\clearpage

\section*{Extended Data}

\begin{figure}
    \centering
    \includegraphics[width=\textwidth,clip]{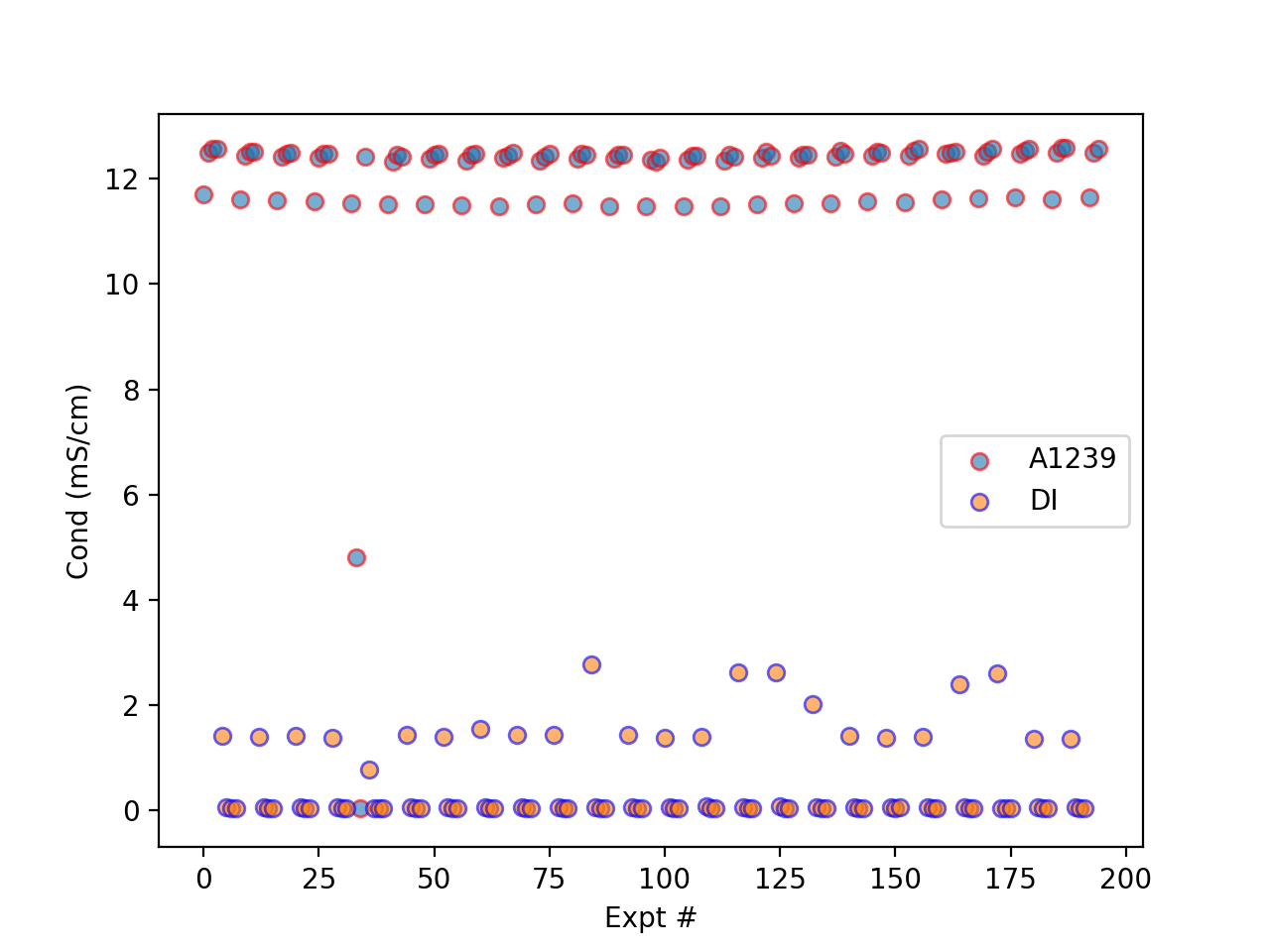}
    \captionsetup{labelformat=empty}
    \caption{\textbf{Extended Figure 1}: Contamination study of extended operation of Clio alternative between conductivity standard (here, acetonitrile and $\ce{LiClO4}$ with conductivity of 12.39 mS/cm; and de-ionized water). Running experiments in triplicate ensures two uncontaminated evaluations per electrolyte.}
    \label{fig:my_label}
\end{figure}

\begin{figure}
    \centering
    \includegraphics[width=\textwidth,clip]{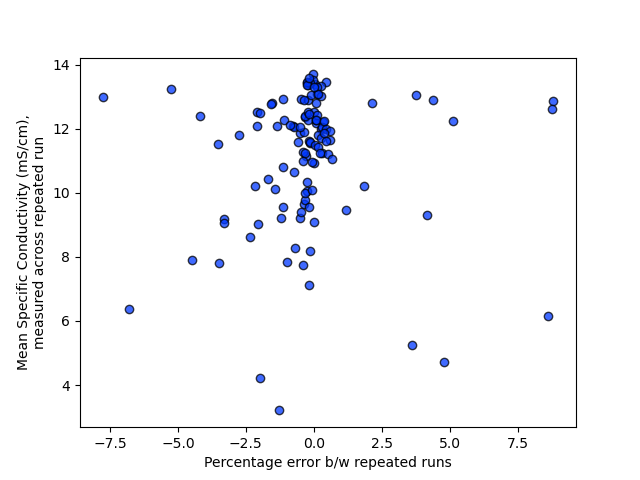}
    \captionsetup{labelformat=empty}
    \caption{\textbf{Extended Figure 2}: Percentage error between repeated evaluations of the same electrolyte, for a range of 120 electrolytes within the EC-DMC-EMC-DEC + $\ce{LiPF6}$ design space. Error is uncorrelated with conductivity level}
    \label{fig:my_label}
\end{figure}

\begin{figure}
    \centering
    \includegraphics[width=\textwidth,clip]{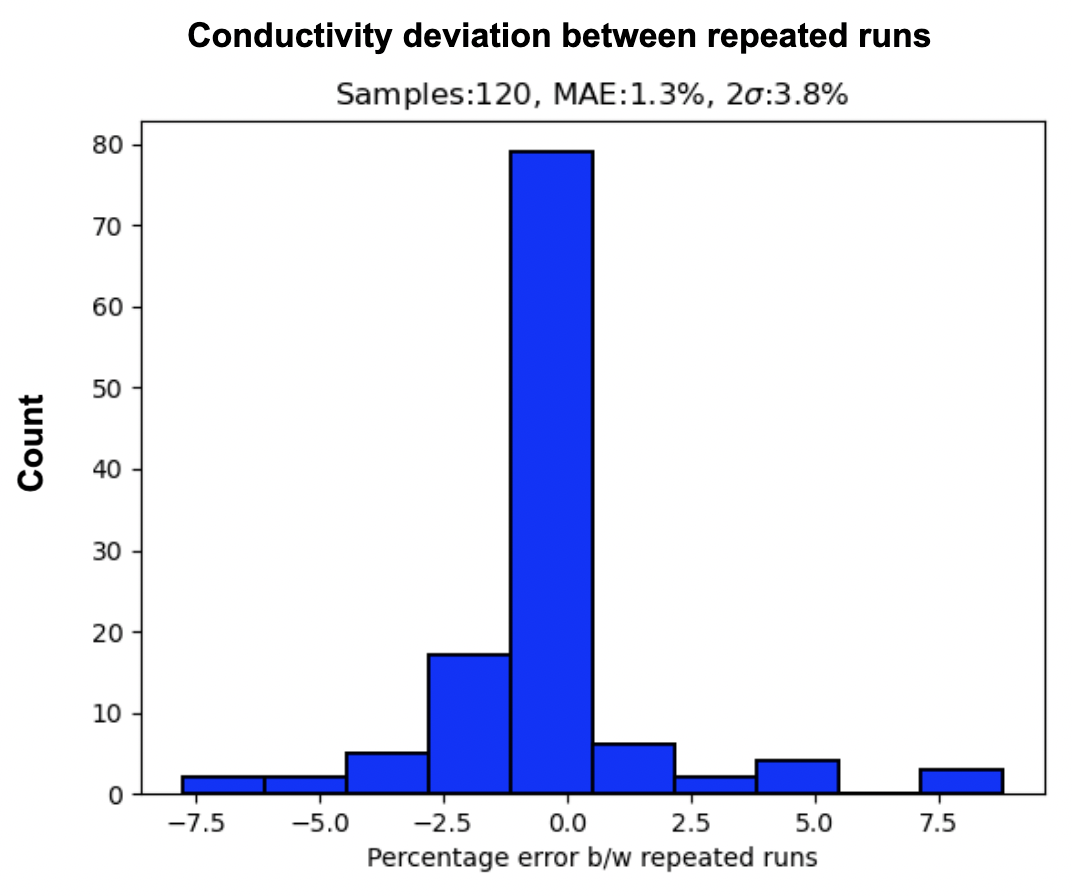}
    \captionsetup{labelformat=empty}
    \caption{\textbf{Extended Figure 3}: Percentage error between repeated evaluations of the same electrolyte, for a range of 120 electrolytes within the EC-DMC-EMC-DEC + $\ce{LiPF6}$ design space. Mean absolute percentage error is 1.3\%, with a 95\% confidence interval of 3.8\% error.}
    \label{fig:my_label}
\end{figure}

\begin{figure}
    \centering
    \includegraphics[width=\textwidth,clip]{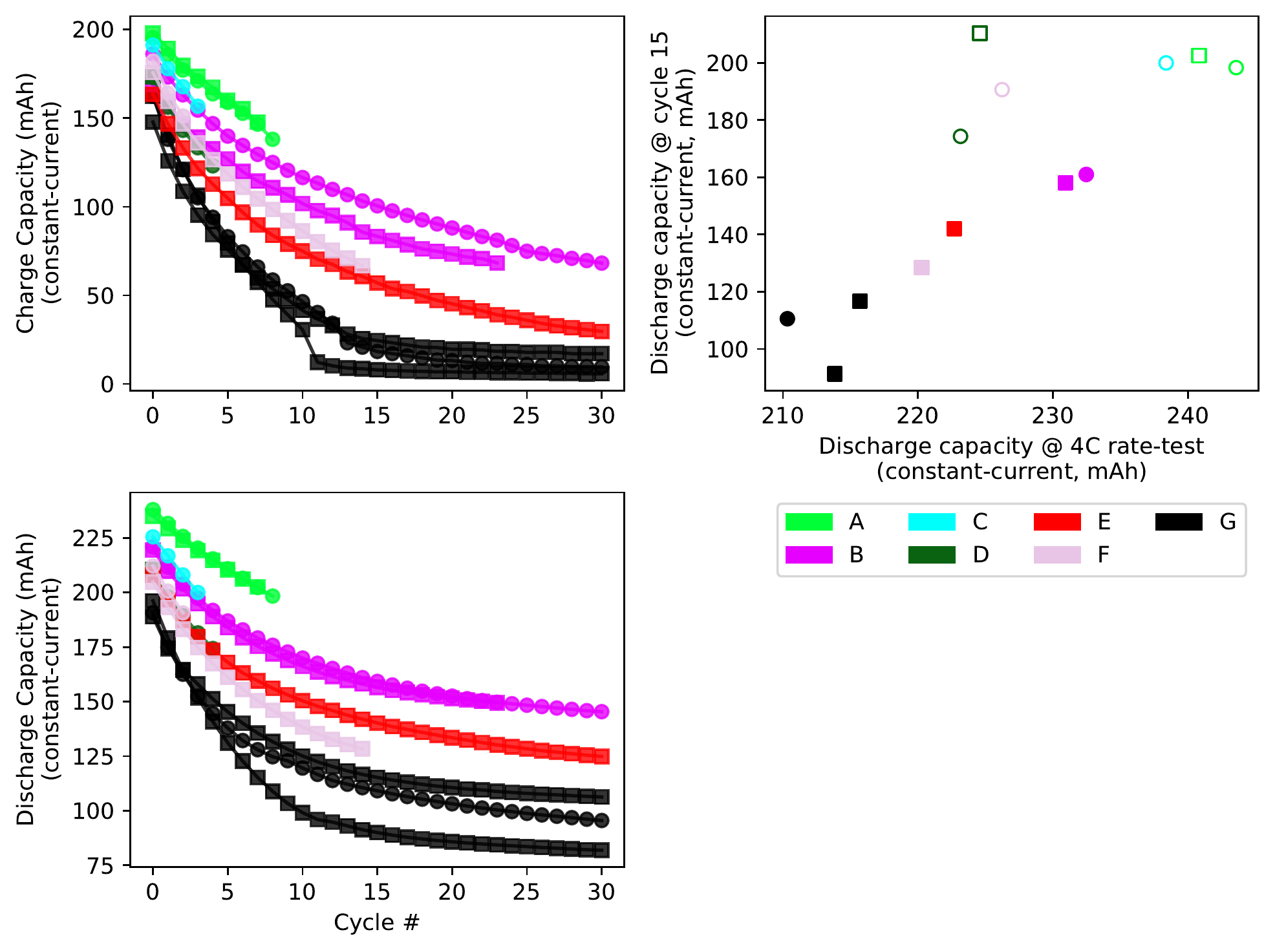}
    \captionsetup{labelformat=empty}
    \caption{\textbf{Extended Figure 4}: Cycling data for each cell during repeated 4C charges, 0.5C discharges over 30 cycles. Only two cells discovered by Clio survive without over-voltage or sudden death to the end - these are the high salt concentration electrolytes tested (E and B).}
    \label{fig:my_label}
\end{figure}

\begin{figure}
    \centering
    \includegraphics[width=\textwidth,clip]{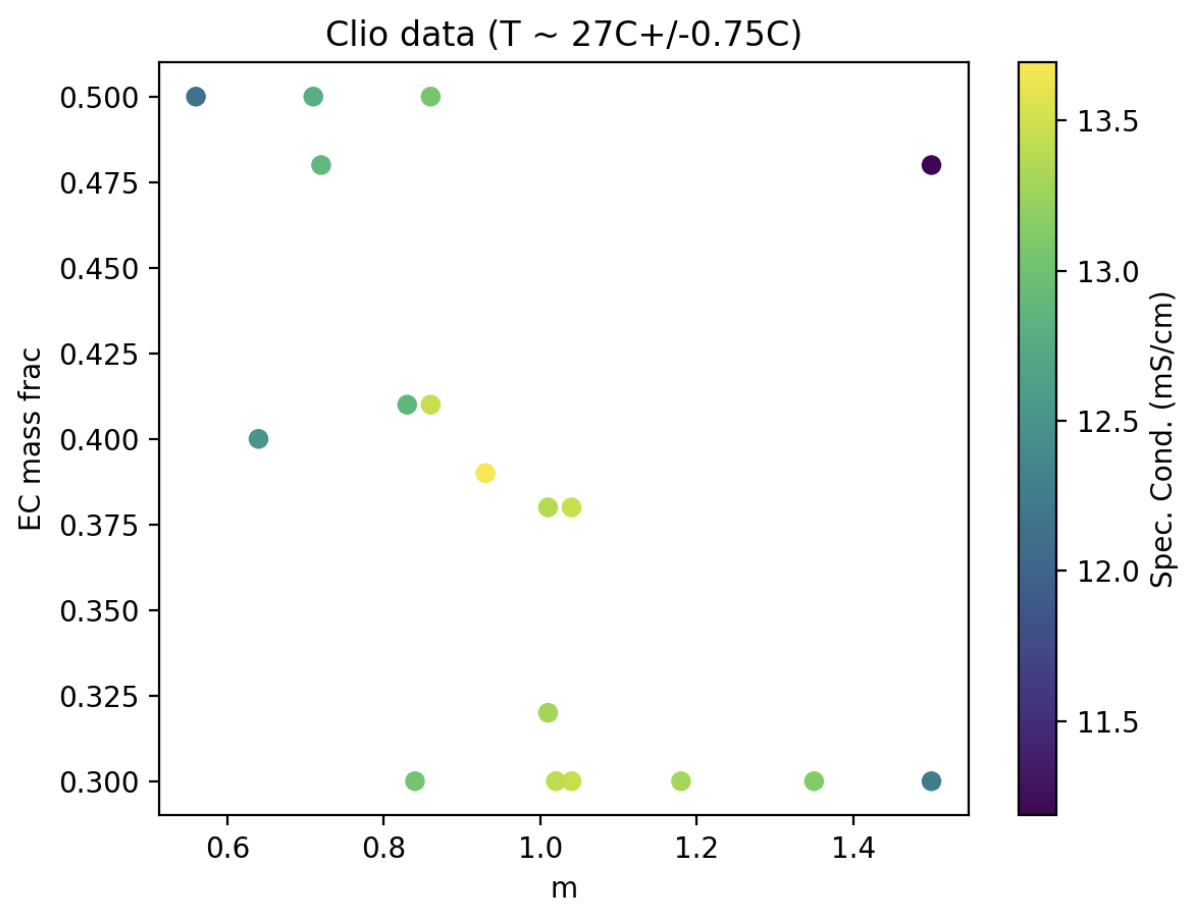}
    \captionsetup{labelformat=empty}
    \caption{\textbf{Extended Figure 5}: EC:DMC only, \ce{LiPF6} electrolytes sampled during optimization (i.e., the top face of Figure 3 left). The experiments indicate a maximum of conductivity at 26-28\textdegree C near 40\% EC mass fraction.}
    \label{fig:my_label}
\end{figure}

\begin{figure}
    \centering
    \includegraphics[width=\textwidth,clip]{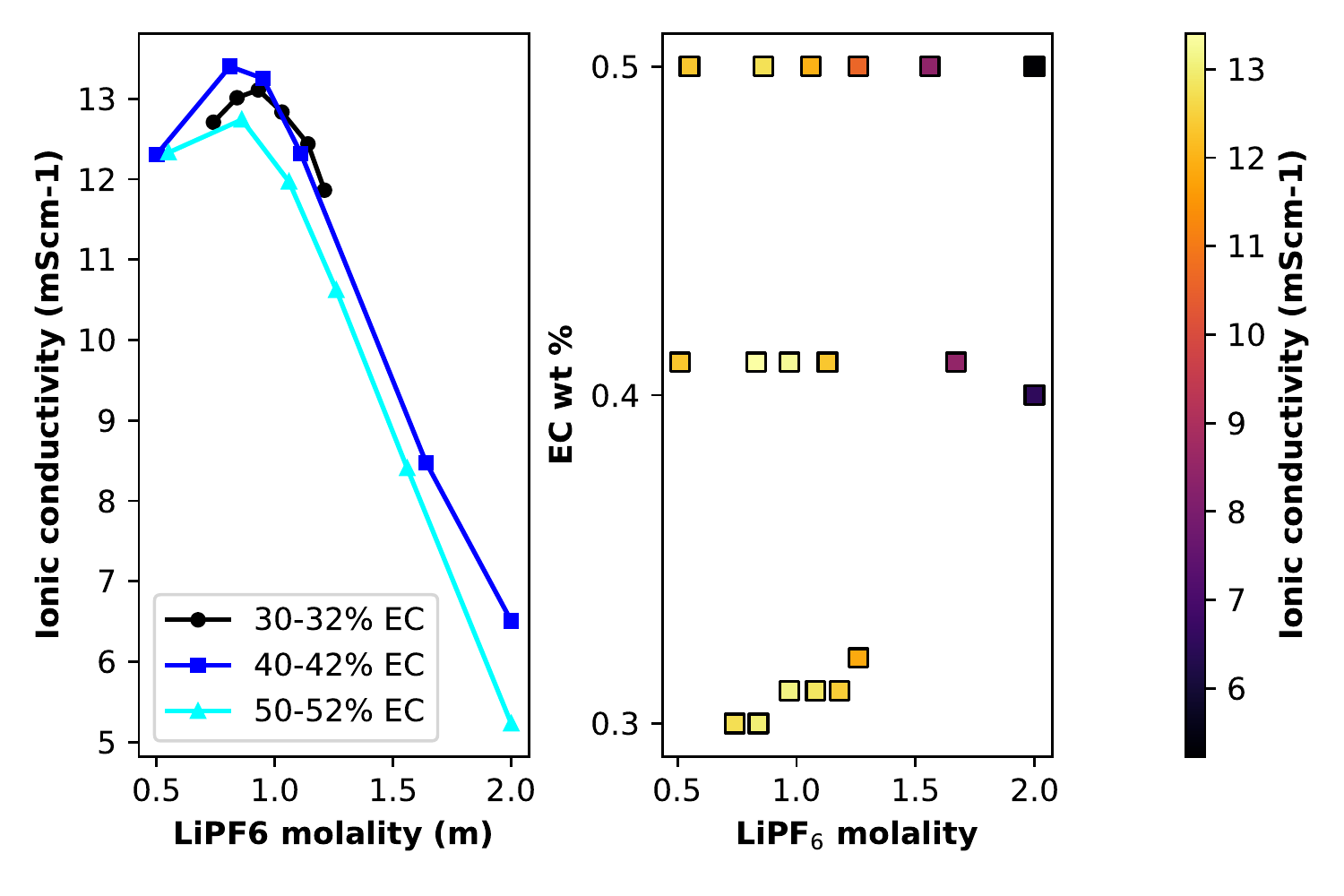}
    \captionsetup{labelformat=empty}
    \caption{\textbf{Extended Figure 6}: Re-surveyed contours on the EC:DMC face of Figure 2 left - due to a small error in density calculation, contours run were within 2\% of each EC mass fraction level. These data show that the maximum of conductivity at 26-28\textdegree C again apppears to be near 40\% EC mass fraction rather than 30\%.}
    \label{fig:my_label}
\end{figure}

\begin{figure}
    \centering
    \includegraphics[width=\textwidth,clip]{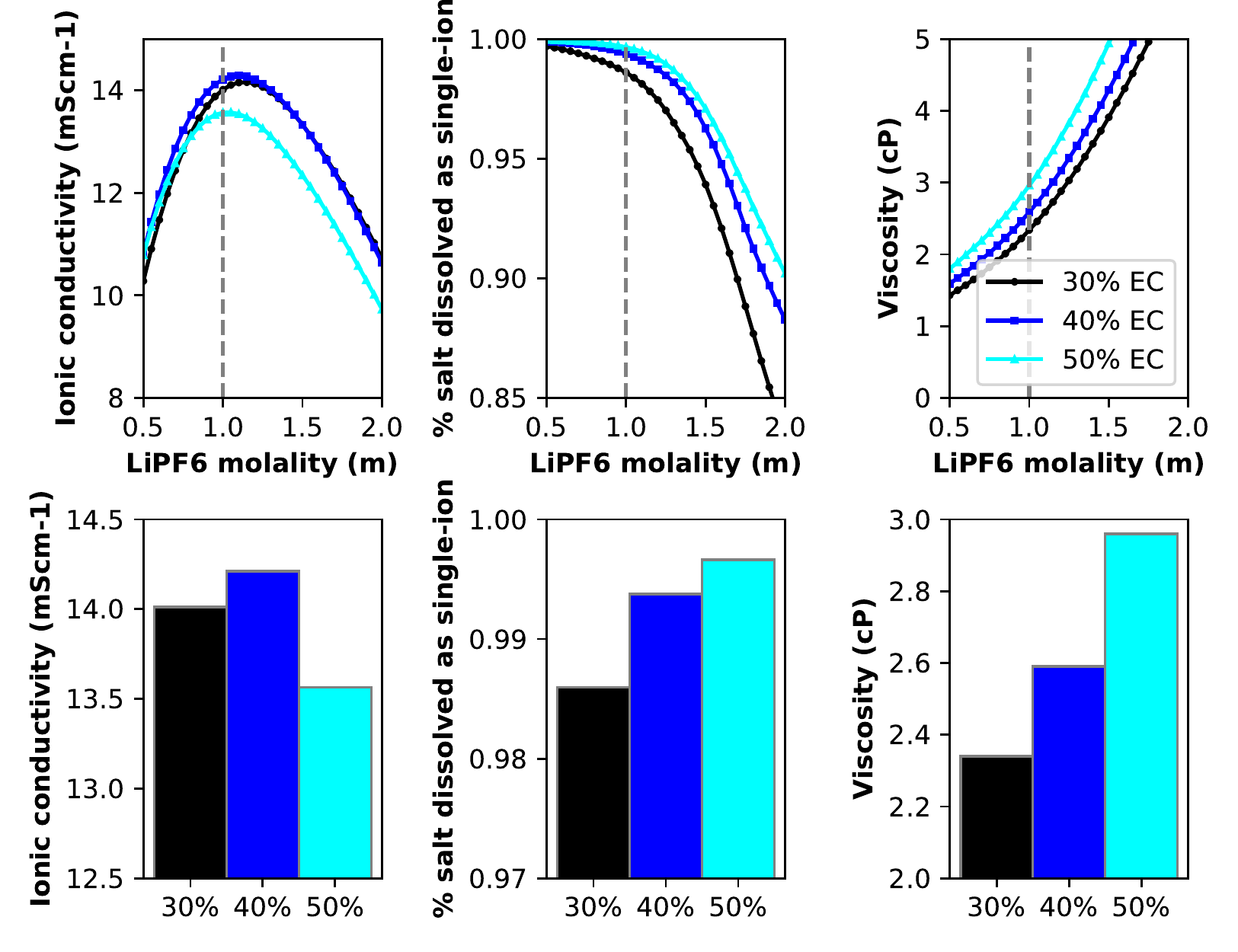}
    \captionsetup{labelformat=empty}
    \caption{\textbf{Extended Figure 7}: Conductivity and two covariates obtained from Advanced Electrolyte Model for EC:DMC \ce{LiPF6} at 30\textdegree C. The model corroborates findings that the maximum conductivity in the system at this temperature is near 40\% EC, though experiments find a higher difference in peak conductivity at a lower molality. Two explanatory covariates - percentage of salt dissolved as single-ions and viscosity - are posited. At 1m \ce{LiPF6}, 40\% EC mass fraction shows single-ion population similar to 50\% EC, but a viscosity more similar to 30\% EC, potentially leading to the increased conductivity.}
    \label{fig:my_label}
\end{figure}

\end{document}